\documentclass[journal,12pt,onecolumn,draftclsnofoot]{IEEEtran}
\usepackage{graphicx}
\usepackage{amsmath}
\usepackage[T1]{fontenc}
\usepackage[utf8]{inputenc}
\usepackage{amssymb,amsfonts} 
\usepackage{url}

\usepackage{theorem}
\theoremstyle{break}
{\theorembodyfont{\itshape} }
{\theorembodyfont{\itshape} \newtheorem{The}{Theorem}}

\newcommand{\lset}{\bigl\{ }
\newcommand{\rset}{\bigr\} }
\newcommand{\lrset}[1]{\lset #1 \rset}

\newcommand{\SOO}{{\mathrm{SO}\left(2\right)}}
\newcommand{\SUG}{{\mathrm{SU}\left(1,1\right)}}

\newcommand{\Complex}{{\mathbb C}}
\newcommand{\Integer}{{\mathbb Z}}

\newcommand{\UDisc}{{\mathcal D}}

\newcommand{\Gsymb}{{\mathrm G}}

\newcommand{\Gelem}[1]{{\mathbf{{#1}}}}
\newcommand{\GGelem}[1]{{\mathit{{#1}}}}
\newcommand{\Gop}[2]{{#1}\left<{#2}\right>}

\newcommand{\absval}[1]{{\left|{#1}\right|}}

\DeclareMathOperator{\shfunction}{sinh}
\DeclareMathOperator{\chfunction}{cosh}
\DeclareMathOperator{\thfunction}{tanh}
\renewcommand{\sinh}{\shfunction }
\renewcommand{\cosh}{\chfunction }
\renewcommand{\tanh}{\thfunction }
\newcommand{\kfun}[1]{{\mathrm{{k}}\left(#1\right)}}
\newcommand{\kfunz}{\mathrm{k}}

\DeclareMathOperator{\efunction}{e}
\newcommand{\expo}[1]{{\efunction^{#1}}}
\DeclareMathOperator{\arctanhfunction}{arctanh}
\newcommand{\arctanh}[1]{{\arctanhfunction\left({#1}\right)}}
\newcommand{\ALFFF}[3]{{\mathfrak P}^{#1}_{#2}\left(#3\right)}
\newcommand{\SConFun}{{\mathfrak P_{-1/2+i\kappa}\left(\cosh \tau\right)}}
\newcommand{\AConFun}[1]{{\mathfrak P_{-1/2+i\kappa}\left(#1\right)}}

\title{The Mehler-Fock Transform and some Applications in  Texture Analysis and Color Processing}
\author{Reiner Lenz
\thanks{R. Lenz is with the Department of Science and Technology, Linköping University, SE-60174 Norrköping, Sweden. The support of the Swedish Research Council through a framework grant for the project Energy Minimization for Computational Cameras (2014-6227) and by the Swedish Foundation for Strategic Research through grant IIS11-0081 (Datadriven scenkaraktärisering för realistisk rendering) is gratefully acknowledged.}}

\begin{document}
%
\maketitle
\begin{abstract}
Many stochastic processes are defined on special geometrical objects like spheres and cones. We describe how tools from harmonic analysis, i.e. Fourier analysis on groups, can be used to investigate probability density functions (pdfs) on groups and homogeneous spaces. We consider the special case of the Lorentz group SU(1,1) and the unit disk with its hyperbolic geometry, but the procedure can be generalized to a much wider class of Lie-groups. We mainly concentrate on the Mehler-Fock transform which is the radial part of the Fourier transform on the disk. Some of the characteristic features of this transform are the relation to group-convolutions, the isometry between signal and transform space, the relation to the Laplace-Beltrami operator and the relation to group representation theory. We will give an overview over these properties and their applications in signal processing. We will illustrate the theory with two examples from low-level vision and color image processing.
\end{abstract}
%
%
\section{Introduction}
\label{sec:intro}
In many applications we know that the data or measurement vectors do not fill the whole vector space but that they are confined to some special regions. Often the form of these regions is defined by problem specific constraints. Typical examples are color vectors. The values in these vectors represent photon counts and therefore they must be non-negative and color vectors are all constrained to lie in the positive part of the vector space. Apart from the measurement vectors we are also interested in transformations acting on these measurements. A typical example in the case of color is the multiplication of the vectors with a global constant. This can describe a global change of the intensity of the illumination or a simple change in the measurement units. It can be shown that many non-negative measurement spaces can be analyzed with the help of Lorentz groups acting on conical spaces. Details of the mathematical background and investigations of spectral color databases can be found in~\cite{Lenz2005,Lenz2008}. Similar geometric structures are used in perception-based color spaces such as the standard CIELAB color system. Here the L-part (related to intensity) is given by the positive half-axis. The (a,b)-part describing the chromatic properties of color are often parameterized by polar coordinates where the radial part corresponds to saturation (given by a finite interval) and the angular part represents hue. A second example that will be considered comes from low-level signal processing. Basic edge-detectors for two-dimensional images come in pairs and the resulting two-dimensional result vectors are often transformed into polar coordinates where the radial part describes the edge-magnitude and the angular part the edge orientation. In many machine learning approaches the raw magnitude value is further processed by an activation function which maps the non-negative raw output to a finite value. A typical activation function is the hyperbolic tangent function~$\tanh$  and we see that output of these edge detector systems are located on the unit disk.  

In the following we describe how harmonic analysis (the generalization of Fourier analysis to groups) can be used to investigate functions defined on special domains with a group theoretical structure. We will present the following results: the investigated functions are defined on a domain with a group of transformations acting on this domain and in this case they can be interpreted as functions on the transformation group. The transformation group defines an (essentially unique) transform that shares many essential properties of the ordinary Fourier transform. Examples of such properties are the simple behaviour under convolutions, preservation of scalar products in the orignal and the transform domain and the close relation to the Laplace operator. This will be illustrated for the unit disk and its transformation group~$\SUG$. The functions corresponding to the complex exponentials are in this case the associated Legendre functions and the (radial) Fourier Transform on the unit disk is the Mehler-Fock transform (MFT). The associated Legendre functions are also eigenfunctions of the hyperbolic Laplace operator and the MFT can therefore be used to study properties related to the heat equation on the unit disk. As mentioned before we will illustrate the theory with some examples from color processing and low-level signal processing. We will mainly study probability distributions on the unit disk and show how the MFT can be used to investigate textures and to manipulate color properties of images. 

Probability distributions on manifolds and their kernel density estimators have been studied before (see, for example, \cite{Pelletier2005}, \cite{Asta2015},\cite{Bates2014}, \cite{Subbarao2009} \cite{Henry2009} and~\cite{Chevallier2015}). Often they are assuming that the manifold is compact. Here we study functions on the unit disk as the simplest example of a non-compact manifold but apart from the Riemannian geometry it also has a related group structure which is crucial in the derivation of the method. The MFT is a well-known tool in mathematics and theoretical physics but there are very few studies relevant for signal processing. Some examples are described in~\cite{Terras1985} and~\cite{Chevallier2015} describes an application of hyperbolic geometry in radar processing. An application to inverse problems is described in~\cite{Huckemann2010}. To our knowledge this is the first application to color signal processing. 

The structure of the paper is as follows: First we introduce the relevant group~$\SUG$ and describe how it operates on the unit disk. We also collect some basic facts of the disk-version of hyperbolic geometry and introduce the parametrization of the unit disk by elements of the factor group~$\SUG/\SOO$. Next we introduce some facts from the representation theory of the group $\SUG$ and its relation to the associated Legendre functions. Restriction to zonal functions (that are functions of the radius only) leads to the MFT which is a central result in the harmonic analysis of this group. The theoretical concepts will be illustrated with two examples, one from color science and the other from texture analysis. For the color image we show how the MFT can be used to desaturate local color distributions and for the textures images we show how it can be used to characterize the textures in the normalized Brodatz texture database described in~\cite{Abdelmounaime2013}. No claim is made regarding the performance of the method compared with other methods in color or texture processing.  
  
\section{Kernel Density Estimators on a Hyperbolic Space}
\label{sec:KDE}

In the following we consider points on the unit disk as complex variables~$z$ and we introduce the group~$\SUG$ consisting of the $2\times 2$ matrices with complex elements:
  \begin{equation}\label{eq:groups}
     \SUG = \lrset{\Gelem{M} = \begin{pmatrix}
   a & b \\
   \overline{b} & \overline{a} \\
 \end{pmatrix}, a,b\in\Complex, \absval{a}^2 - \absval{b}^2 = 1}
  \end{equation}
The group operation is the usual matrix multiplication and the group acts as a transformation group on the open unit disk~$\UDisc$ (consisting of all points~$z\in\Complex$ with~$\absval{z}<1$) as the Möbius transforms:
\begin{equation}\label{eq:su11transf}
   \left(\Gelem{M},z\right)
   =
   \left(\begin{pmatrix}
   a & b \\
   \overline{b} & \overline{a}
 \end{pmatrix},z \right)
 \mapsto
 \Gop{\Gelem{M}}{z}=\frac{az+b}{\overline{b}z+\overline{a}}, z\in\UDisc
\end{equation}
The concatenation of two transforms correspond to the multiplication of the two corresponding matrices which gives: $\Gop{\left(\Gelem{M}_1\Gelem{M}_2\right)}{z} = \Gelem{M}_1\left<\Gelem{M}_2\left<z\right>\right>$ for all matrices and all points. In the following we will use the notation~$\Gelem{M}$ if we think of the group elements as matrices. If we want to stress their role as elements of a group or if we want to parametrize the Möbius transforms then we will often use the symbol~$\GGelem{g}$.  Every three-dimensional rotation can be written as a product of three rotations around the coordinate axes.  A similar decomposition holds for~$\SUG$ whose elements are:
\begin{equation}\label{eq:CD}
\GGelem{g}(\varphi,\tau,\psi) = \begin{pmatrix}
   \cosh \frac{\tau}{2} \expo{i(\varphi + \psi)/2}& \sinh \frac{\tau}{2}\expo{i(\varphi - \psi)/2} \\
   \sinh \frac{\tau}{2}\expo{-i(\varphi - \psi)/2} & \cosh \frac{\tau}{2}\expo{-i(\varphi + \psi)/2} \\
 \end{pmatrix}
\end{equation}
with three parameters~$\psi, \tau, \varphi.$ 

Defining subgroups~$K = \lrset{ \GGelem{g}(\varphi,0,0)}$ and~$A \lrset{\GGelem{g}(0,\tau,0)}$ we write the decomposition as~$\SUG = KAK.$  This is known as the Cartan or polar decomposition of the group and~$\psi, \tau, \varphi$ are the Cartan coordinates. 

Matrices in~$K$ are rotations, leaving the origin fixed and applying a general element in~$\SUG$ to the origin~$0$ gives: 
\[
\Gop{\GGelem{g}(\varphi,\tau,\psi)}{0} =  \tanh\frac{\tau}{2}\expo{i\varphi}.
\]
Defining two transformations as equivalent if the parameters~$\varphi,\tau$ are identical, we obtain a correspondence between the points on the unit disk and the equivalence classes of transformations. We write~$\UDisc = \SUG/K$ and functions on the unit disk are functions on~$\SUG$ that are independent of the last argument of the Cartan decomposition.

For two points~$w,z\in\UDisc$ we define the kernel function~$\kfun{w,z}$ as the probability of obtaining the true measurement~$w$ given that the actual measurement resulted in the point~$z$. Often it is natural to require
\[
\kfun{\Gop{\Gelem{M}}{w}, \Gop{\Gelem{M}}{z}} = \kfun{w,z}, \text{ for all } \Gelem{M}\in\SUG
\]
for example in the case of invariance against coordinate system changes. The Möbius transforms define an invariant, hyperbolic, distance between points~$z$ and~$w,$ given by
\[
d(w,z) = 2\arctanh{\absval{\dfrac{z-w}{1-\overline{z}w}}}
\]
and we consider only kernel functions of the form~$\kfun{d(w,z)}$. Points on the unit disk are equivalence classes of group elements and in the general case we consider ((left-isotropic) kernels~$\kfun{\GGelem{g},\GGelem{g}_l}$ on the full group satisfying 
\[
\kfun{\GGelem{g},\GGelem{g}_l} = \kfun{\GGelem{hg},\GGelem{hg}_l} \text{ for all }\GGelem{g, g}_l, \GGelem{h}\in\SUG.
\]
It follows that~$\kfun{\GGelem{g,g}_l} = \kfun{\GGelem{g}_l^{-1}\GGelem{g,e}} = \kfun{\GGelem{g}_l^{-1}\GGelem{g}}$ where~$\GGelem{e}$ is the identity element. In the case of the unit disk we consider elements~$\GGelem{g}_l  =  g(\varphi_l,\tau_l,0)$ and~$\GGelem{g}=g(\varphi_0,\tau_0,0)$ and we find the relation between these two decompositions and the difference~$\GGelem{g}_l^{-1}\GGelem{g} = g(\varphi,\tau,0)$ as (see~\cite{Vilenkin2012}, (Vol 1, page 271)):
\begin{equation}
\label{eq-ttrans}
  \cosh\tau =  \cosh\tau_l\cosh\tau_0 + \sinh\tau_l\sinh\tau_0\cos(\varphi_l-\varphi_0)
\end{equation}
We therefore consider only radial kernels of the form:
\[
\kfun{\GGelem{g,g}_l} = \kfun{\cosh\tau_l\cosh\tau_0+\sinh\tau_l\sinh\tau_0\cos(\varphi_l-\varphi_0)}.
\]

In general we have to compute the function values~$\kfun{\GGelem{g,g}_l}$ for every pair~$(\GGelem{g},\GGelem{g}_l).$ Of special interest are functions which separate these factors. They are the associated Legendre functions (zonal or Mehler functions, \cite{Vilenkin2012}, page~324) of order~$m$ and degree~$\alpha = -1/2+i\kappa$ defined as (see Eq.(\ref{Eq-ALfunDef})):
\begin{equation}\label{Eq-ALfunDef}
    \ALFFF{m}{\alpha}{\cosh\tau}\!=\!\frac{\Gamma(\alpha+m+1)}{2\pi\Gamma(\alpha+1)}\!\int_0^{2\pi}\hspace*{-2mm}\left(\sinh\tau\cos\theta +\cosh\tau\right)^{\alpha}\expo{im\theta}\;d\theta
\end{equation}
with the addition formula (\cite{Vilenkin2012}, page~327) 
\begin{eqnarray}
\label{eq:addform}
\,&\ & \hspace*{-2cm}\ALFFF{\ }{\alpha}{\cosh\tau_l\cosh\tau_0 + \sinh\tau_l\sinh\tau_0\cos\theta}\nonumber\\
&=&\sum_{m\in\Integer} \ALFFF{-m}{\alpha}{\cosh\tau_l}\ALFFF{m}{\alpha}{\cosh\tau_0}\expo{-im\theta}
\end{eqnarray}
where ${\mathfrak P}^{0}_{\alpha} = {\mathfrak P}_{\alpha}$. It seperates dependent terms, (from data~$g_l$ with parameters~$\tau_l,\varphi_l$) and the the kernel part (from~$g$ with~$\tau_0,\varphi_0$).

This is generalized by the MFT, showing that a large class of functions are combinations of the associated Legendre functions:
\begin{The}[Mehler-Fock Transform; MFT]
\label{The:MFT}
For a function~$\kfunz$ defined on the interval~$\left[1,\infty\right)$ define its transform~$c$ as:
\begin{equation}\label{MFT-Cos}
    c(\kappa) = \int_0^\infty \kfun{\cosh\tau}\SConFun\sinh\tau\;d\tau
\end{equation}
Then $\kfunz$ can be recovered by the inverse transform:
\begin{equation}\label{IMFT}
    \kfun{\cosh\tau} = \int_0^\infty \kappa \tanh(\pi\kappa)\AConFun{\cosh\tau} c(\kappa)\;d\kappa
\end{equation}
\end{The}
Details about the transform, special cases and its applications can be found in~\cite{Terras1985},\cite{Vilenkin2012},\cite{Lebedev1972},\cite{Sneddon1972} and~\cite{Project1954a}.

In the general group theoretical context the integral in Eq.(\ref{MFT-Cos}) defines a convolution over the subgroup~$A$ parametrized by the hyperbolic angle~$\tau.$ In the case of the group~$\Gsymb = \SUG$ this general construction can be applied by using the Cartan coordinates of group elements. The analysis of the angular parts lead to ordinary Fourier series and for the radial part we find for the MFT (in hyperbolic coordinates) the integrals (using Eq.(\ref{eq-ttrans}) characterizing the contribution from datapoint~$l$:
\begin{eqnarray}
& &\hspace*{-1cm} \int_0^\infty \kfun{\cosh \tau}\AConFun{\cosh\tau_l\cosh\tau + \sinh\tau_l\sinh\tau\cos\theta}\sinh\tau\;d\tau\nonumber\\
&=&\sum_{m\in\Integer}\expo{-im\theta}\ALFFF{-m}{1/2+i\kappa}{\cosh\tau_l}w_{\kappa m}^{\left(k\right)} = \sum_{m\in\Integer}\gamma_{\kappa ml}^{\left(d\right)}w_{\kappa m}^{\left(k\right)}
\label{eq:hacoef}
\end{eqnarray}
with~$w_{\kappa m}^{\left(k\right)} = \int_0^\infty\kfun{\cosh \tau} \ALFFF{m}{1/2+i\kappa}{\cosh\tau}\sinh\tau\;d\tau$. The coefficients~$\gamma_{\kappa ml}^{\left(d\right)}$ are computed from the measurements and the weights $w_{\kappa m}^{\left(k\right)}$ are independent of the data. 

It is also known that the MFT preserves the scalar product (Parseval relation). If~$c_n(\kappa)=\int_1^\infty f_n(x)\ALFFF{\ }{1/2+i\kappa}{x}\;dx$  then 
\begin{equation}
\int_0^\infty c_1(\kappa)c_2(\kappa) \kappa \tanh(\pi \kappa) \;d\kappa =
\int_1^\infty f_1(x)f_2(x)\;dx
\label{eq:Parseval}
\end{equation}
This can be found in~\cite{Sneddon1972}(7.6.16) and~\cite{Sneddon1972}(7.7.1) where the interested reader can find more information on the MFT. 
Finally we mention that the conical functions are eigenfunctions of the Laplace-operator which is the operator that commutes with the group action. The eigenvalues are~$-(1/4+\kappa^2)$ (see~\cite{Dym1972} Chap. 4 for a brief summary of these results). 

\section{Illustrations}
\label{sec:illustrations}

\begin{figure}[htb]
\begin{minipage}[b]{1.0\linewidth}
  \centering  
	\centerline{\includegraphics[width=0.95\textwidth]{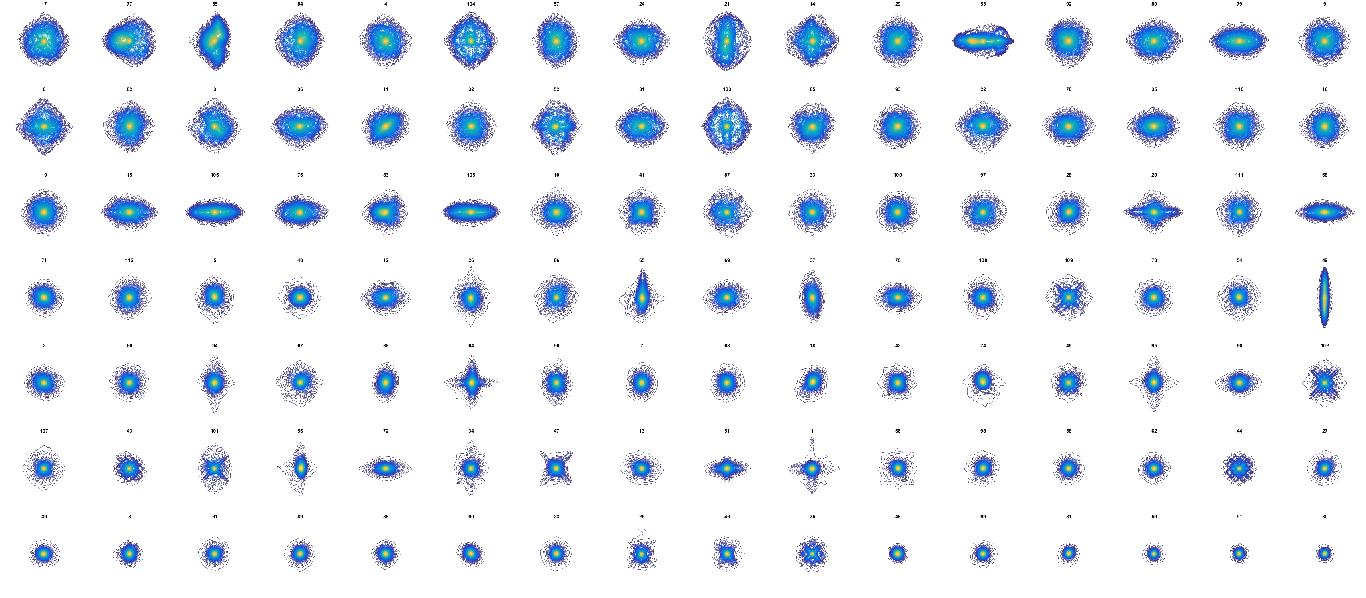}}
  \centerline{(a) Sorted densities}\medskip
\end{minipage}
\begin{minipage}[b]{1.0\linewidth}
  \centering  
\centerline{\includegraphics[width=0.95\textwidth]{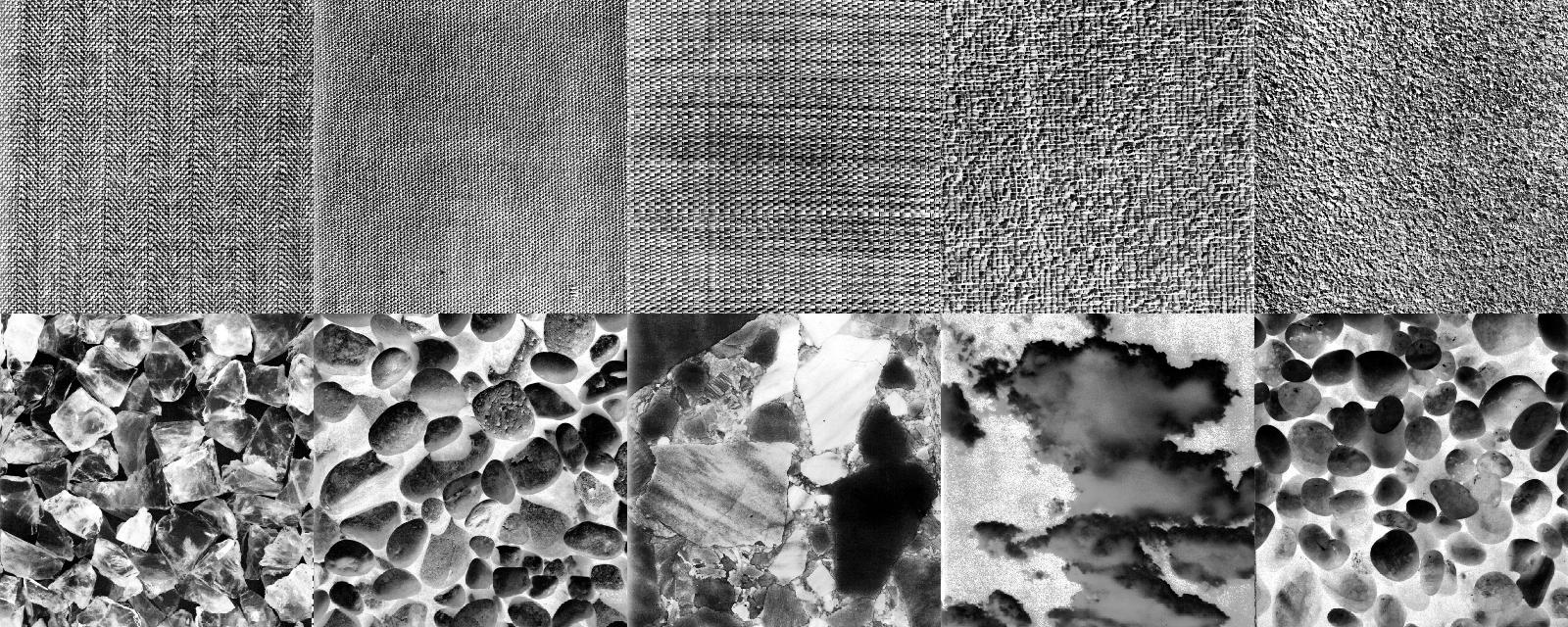}}
  \centerline{(b) First and last textures in the sequence}\medskip
\end{minipage}
\caption{Texture Analysis.}
\label{fig:Brodatz}
\end{figure}

In the first example we use the MFT to study a simple filter system (more information can be found in~\cite{Lenz1995}). The input of the system are the pixel values on a four-point orbit of the dihedral group, for example the four corner points of a square. The theory of group representations for the dihedral group shows that a natural filter pair consists of the edge filters with coefficients~$(1,1,-1,-1), (1,-1,-1,1)$ and the raw output values~$dx,dy$ should be transformed into polar coordinates~$\rho=\sqrt{dx^2+dy^2},\varphi = \arctan(dx,dy).$ They are the familiar edge magnitude (independent of dihedral transformation) and the edge-orientation. Following popular practice in machine learning the raw magnitude result~$\rho$ is converted to a value~$r$ less than or equal to one with the help of a transfer function. Every configuration of a four-pixel distribution results thus in a point on the unit disk. In our first experiment we compute the probability distribution of such a filter system for each of the 112 images in the normalized Brodatz texture database (described in~\cite{Abdelmounaime2013}). The kernels used for the density estimators are of the form~$(\cosh \alpha)^{-s}$ for which the MFT is known. The resulting probability density functions on the disc where then analysed with the help of the MFT and the ordinary Fourier transform in the angular variable. In the following example we used the orientation invariant component of the angular FFT and then computed its MFT-values. Next we applied the Parseval relation (Eq.(\ref{eq:Parseval})) to compute the distance between pairs of MFTs. Multidimensional scaling was then used to project the distance distributions to the two-dimensional plane. Ignoring the orientation information and only using the magnitude values results in a characterization of the textures in terms of overall roughness which is illustrated in Figure~\ref{fig:Brodatz}. It turns out that the first axis in the multidimensional scaling mapping is by far the most dominant and we therefore sort the textures using their position on this first axis. The contour plots of the sorted sequence of these pdfs is shown in Figure~\ref{fig:Brodatz}(a). The first and last five textures in the sequence are shown in Figure~\ref{fig:Brodatz}(b). 

\begin{figure}[htb]	
	\begin{minipage}[b]{1.0\linewidth}
  \centering
  \centerline{\includegraphics[width=0.95\textwidth]{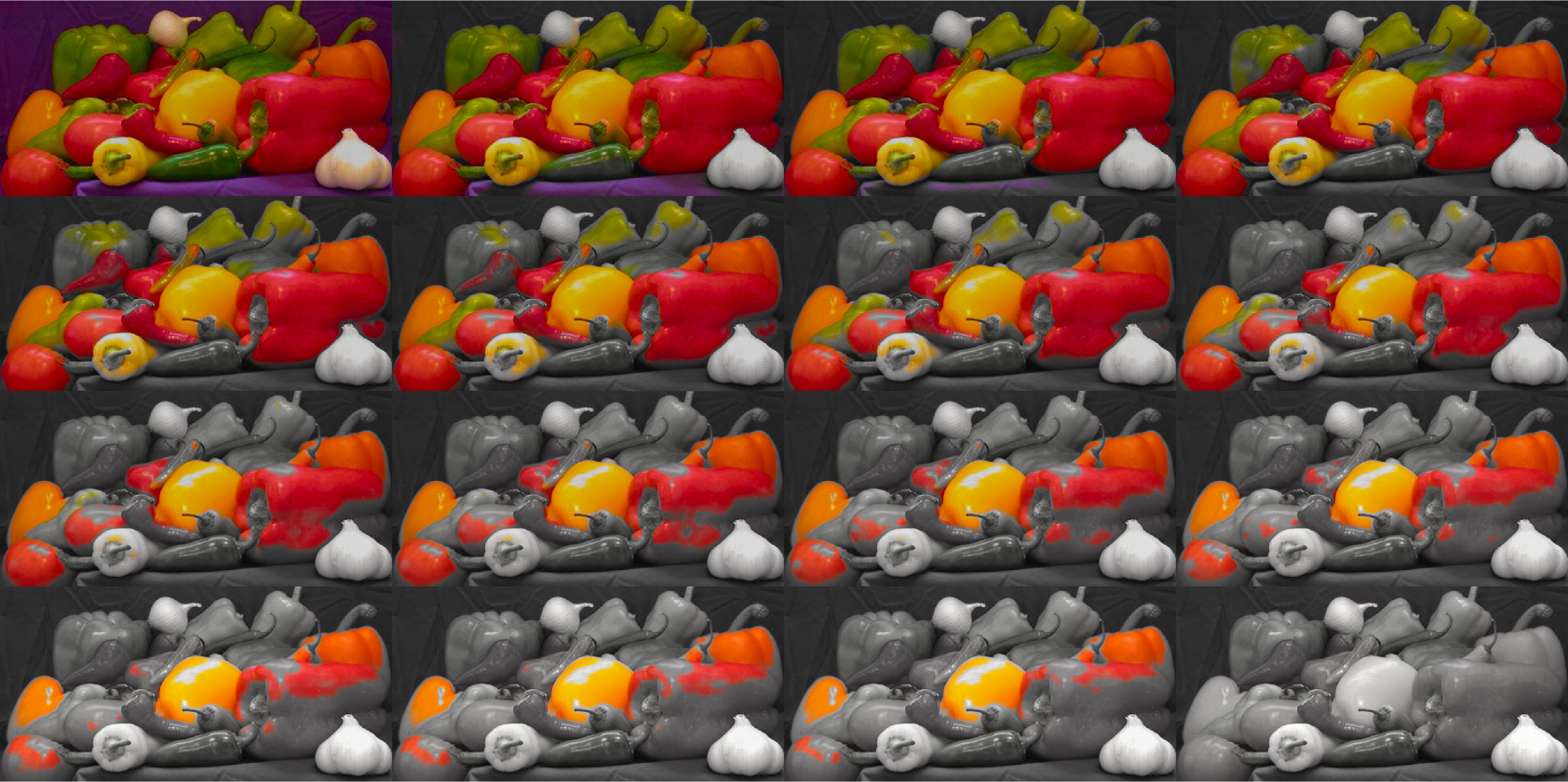}}
\end{minipage}
\caption{Desaturation of a color image.}
\label{fig:pepmon}
\end{figure}

In the color image illustration we convert first the original RGB image into its CIELAB form. The (a,b) part of a CIELAB vector describes the chroma properties of a color and we convert the original (a,b) vector to polar coordinates and scale the radial part so that all points are located on the unit disk. For each point in the image we then select a window (of size $10\times 10$ pixels) and compute the MFT of the distribution of the radial (saturation) variable. From the fact that the conical functions are eigenfunctions of the Laplace operator it follows that we can simulate a kind of heat-flow on these probability distributions of the saturation values. In the MFT space the ``high-frequency'' components with big~$\kappa$ values are faster suppressed due to the negative eigenvalue of the conical functions. In the experiment we computed first the probability distributions of the local saturation values. We then used the MFT to transform these original distributions and then we used a sequence of increasing time-intervals to simulate the dissipation of the distributions. Due to the form of the conical functions we expect that image points with concentrated high saturation values should survive longest whereas low saturation values and inhomogenous distributions should disappear faster. After the operation in the transform domain we transformed the new function back and we detected the mode (saturation with the highest probability value). This mode-saturation value is used as the new saturation at this point, the angular-hue and the intensity value of the CIELAB L-component are copied from the original and the color vectors are transformed back into RGB-form. The result of a sequence of sixteen such time-intervals is shown in Figure~\ref{fig:pepmon}. Here the first fifteen use equal time-increments whereas the last image corresponds to a time-interval that is twice as long as the previous one. 
 
\section{Summary and Conclusions}
\label{sec:SandC}

We observed that many processes produce signals that are located on the unit disk. Apart from the property that the outpout vectors are points on the unit disk it is also important that operations on the disk have a meaningful interpretation. In the case of the edge filter systems these operations are changes in orientation and edge-magnitude. For the chroma-vectors the corresponding operations are hue and saturation changes. The next step is the observation that functions on the unit disk have an interpretation as functions on the Lorentz group~$\SUG.$ For this group there is a (more or less) unique transform with similar properties as the ordinary Fourier transform. This transform is the Mehler-Fock transform and the functions that correspond to the complex exponentials are the conical functions. Based on the properties of the MFT we showed that (a) processing can be done in the original or the transform domain (b) the relation between the MFT and group convolution leads to a separation of the kernel density estimator into data-dependent and kernel-dependent terms (c) the MFT preserves geometry between the original and the transform domain and we can therefore estimate similarity between signals in either space and (d) we can use the eigenfunction property of the Laplacian to study methods for smoothing the data. The detailed study of numerical aspects such as approximation errors, sampling properties, and comparison to other methods is left for future investigations.


\end{document}